\documentclass{article}
\usepackage{spconf,amsmath,graphicx}

\usepackage[colorlinks=true,allcolors=black]{hyperref}
\usepackage{url}
\usepackage{textcomp} 
\usepackage{gensymb} 
\usepackage{arydshln}
\usepackage{booktabs}
\usepackage{multirow}
\usepackage[utf8]{inputenc}
\usepackage[acronym]{glossaries}
\usepackage[caption=false]{subfig}
\usepackage[table,dvipsnames]{xcolor}
\usepackage{xspace}
\usepackage[sort&compress,numbers]{natbib}
\usepackage{balance}

\usepackage{amssymb}
\usepackage{latexsym}
\usepackage{url}
\usepackage{enumitem}

\newacronymstyle{long-short-br}
{%
  \GlsUseAcrEntryDispStyle{long-short}%
}%
{%
  \GlsUseAcrStyleDefs{long-short}%  
}
\setacronymstyle{long-short-br}

\title{\Large Unconstrained Periocular Recognition: Using Generative Deep Learning Frameworks for Attribute Normalization}

\makeatletter
\def\@name{ \emph{Luiz~A.~Zanlorensi\textsuperscript{1,2}, Hugo~Proença\textsuperscript{2,3}, David~Menotti\textsuperscript{1}}\thanks{\scriptsize{\textcopyright 20XX IEEE. Personal use of this material is permitted. Permission from IEEE must be obtained for all other uses, in any current or future media, including reprinting/republishing this material for advertising or promotional purposes, creating new collective works, for resale or redistribution to servers or lists, or reuse of any copyrighted component of this work in other works.}}\\}
\makeatother

\address{\textsuperscript{1}Department of Informatics, Federal University of Paraná, Curitiba, Brazil\\
\textsuperscript{2}Department of Informatics, University of Beira Interior, Covilhã, Portugal\\
\textsuperscript{3}IT: Instituto de Telecomunicações\\[0.5ex]
\normalsize
\textsuperscript{}\textit{\{lazjunior, menotti\}@inf.ufpr.br} \qquad \textsuperscript{}\textit{hugomcp@di.ubi.pt}}

\begin{document}
\ninept
\sloppy
\maketitle
\begin{abstract}
\textit{Ocular biometric systems working in unconstrained environments usually face the problem of small within-class compactness caused by the multiple factors that jointly degrade the quality of the obtained data. In this work, we propose an attribute normalization strategy based on deep learning generative frameworks, that reduces the variability of the samples used in pairwise comparisons, 
without reducing their discriminability.
The proposed method can be seen as a preprocessing step that contributes for data regularization and improves the recognition accuracy, being fully agnostic to the recognition strategy used.
As proof of concept, we consider the ``eyeglasses''  and ``gaze'' factors, comparing the levels of performance of five different recognition methods with/without using the proposed normalization strategy.
Also, we introduce a new dataset for unconstrained periocular recognition, composed of images acquired by mobile devices, particularly suited to perceive the impact of ``wearing eyeglasses'' in recognition effectiveness.
Our experiments were performed in two different datasets, and support the usefulness of our attribute normalization scheme to improve the recognition performance.}
\end{abstract}


\section{Introduction}
\label{sec:introduction}

\glsresetall

The development of ocular biometric systems operating under unconstrained environments is challenging since the collected data (images) may present some problems caused by noise, blur, motion blur, occlusion, eye gaze, off-angle, eyeglasses, contact lenses, makeup, among others.
These problems generate high within-class variability degrading the level of uniqueness of the features extracted from the biometric trait.

With the recently advancement of deep learning techniques, several approaches applying Convolutional Neural Networks (CNN) to periocular recognition have been developed~\cite{luz2018deep, proenca2018prwis, zhao2018critical, diaz2018periocular, zanlorensi2019cross, diaz2019cross}.
An advantage of applications based on deep learning is that unlike the handcrafted features, there is a process of representation learning.
This process can produce feature extractor models invariant for some within-class factors, depending on the image samples present in the training set.
Nevertheless, new approaches are still being developed using handcrafted features and achieving top-ranked results in ocular recognition competitions~\cite{sequeira2017cross, demarsico2017miche, ahmed2017michewinner, zanlorensi2019survey}.
The main advantage of these approaches is the computational cost compared with methods based on deep learning techniques.

\begin{figure}[!ht]
\centering

   	\includegraphics[width=\columnwidth]{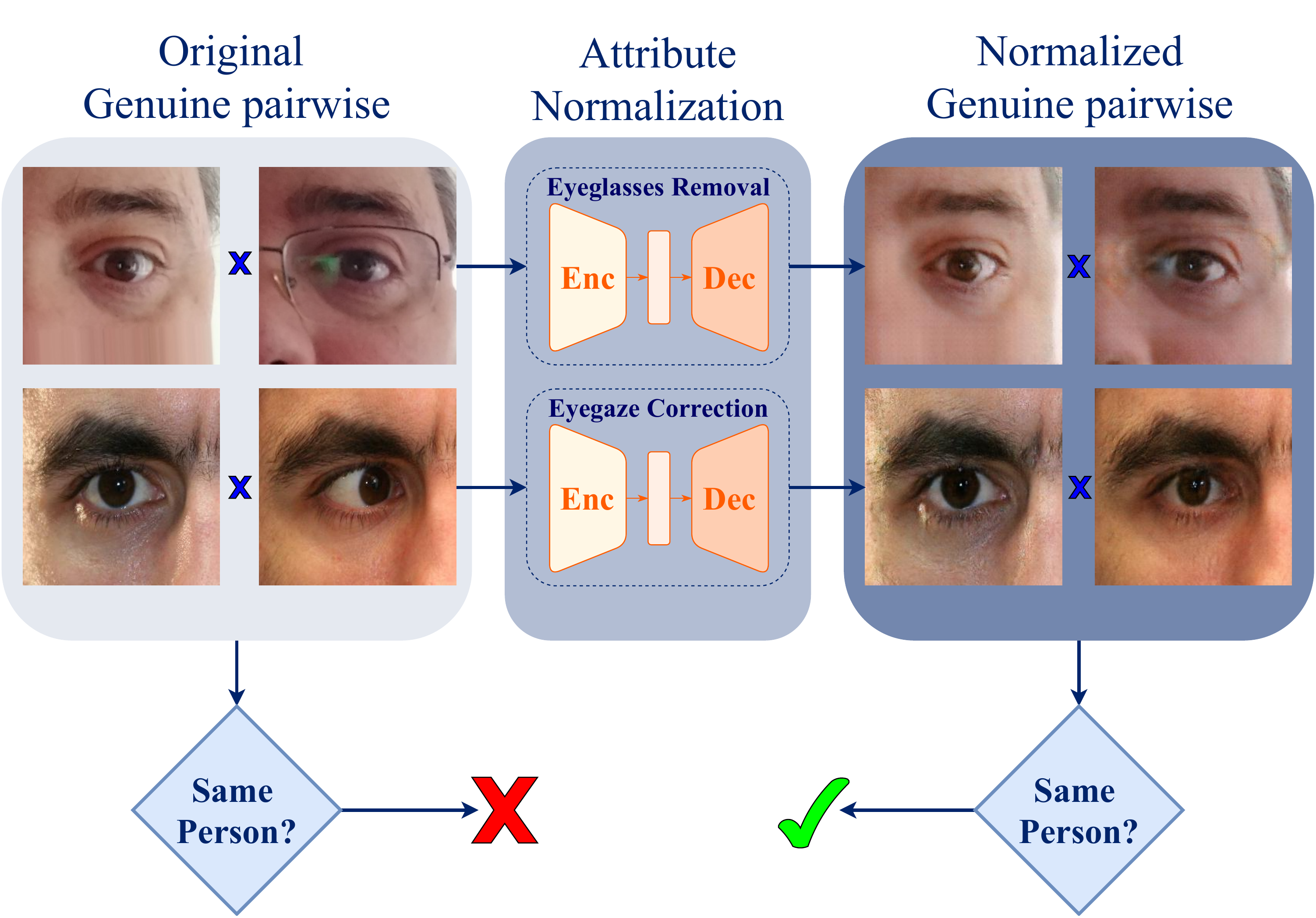}

\caption{Cohesive perspective of the proposed attribute normalization scheme: images feed an encoder/decoder deep model for automatic image editing, removing the eyeglasses and correcting deviated gazes before the recognition step. This contributes for reducing the within-class variability without significantly reducing the discriminability between classes, which is the key for the observed improvements in performance.}
\label{fig:proposed}
\end{figure}

Even though CNN approaches can handle within-class variability, there are still several factors present in images captured under unconstrained environments, which affect periocular recognition in biometric systems based on deep learning and mainly on handcrafted features.
Regarding these kind of problems, our work proposed an image preprocessing method to normalize the most common image attributes that can decrease the recognition effectiveness in periocular biometric systems.
The proposed attribute normalization prepossessing consists of remove or correct attributes that are different in a pairwise image comparison using deep models for image editing, as show in Fig.~\ref{fig:proposed}.
For example, in a dataset containing images from the same subject wearing and not wearing eyeglasses, the proposed preprocess will normalize all the images by removing the eyeglasses.
Another contribution is a new dataset for mobile periocular recognition under a real and slightly constrained environment.
This dataset, called UFPR-Eyeglasses, is composed of images captured by mobile devices from subjects wearing and not wearing eyeglasses.
The rest of this paper is organized as follows.
In Sec.~\ref{sec:related}, we discuss the related works describing deep models for attribute editing.
In Sec.~\ref{sec:proposed}, we explain the proposed normalization and how it was performed.
The experimental protocol is described in Sec.~\ref{sec:experiments} and the results are reported in Sec.~\ref{sec:results}.
Finally, we state conclusions in Sec.~\ref{sec:conclusion}.

\section{Related Work}
\label{sec:related}

Recently, several methods have been developed for automatic facial attribute editing.
Approaches based on Generative Adversarial net (GAN)~\cite{goodfellow2014gan} and Variational Autoencoder (VAE)~\cite{kingma2014vae} architectures reported promising results performing these tasks~\cite{larsen2016vaegan, perarnau2016icgan, choi2018stargan, lample2017fader, he2019attgan, shen2017resface, zhu2017cyclegan, karras2019stylegan, xiao2018elegant}.
The models for face attributes editing can be divide based on their ability to manipulate a single~\cite{shen2017resface, zhu2017cyclegan} or multiple attributes~\cite{larsen2016vaegan, perarnau2016icgan, choi2018stargan, lample2017fader, he2019attgan}, such as eyeglasses, hair color, age, mustache, gender, beard, among others.
Also, there are strategies for image attribute editing by transferring face attributes~\cite{choi2018stargan, xiao2018elegant, karras2019stylegan}.
The concept of this task is to modify a face image based on attributes contained in another image, preserving the subject identity.
As stated by He et al.~\cite{he2019attgan}, one advantage of models based on encoder/decoder architecture is that they can handle multiple attributes manipulation using a single trained model.
Also, in models based on encoder-decoder architecture, the attributes are manipulated through modifications in the latent representation generated by the encoder.
With these modifications, the decoder can generate images with different attributes compared to the input ones.

Regarding the image attribute manipulation, each model proposes a different strategy to relate the latent representation to the face attributes.
The model proposed by Shen and Liu~\cite{shen2017resface} consists of two networks performing the inverse attribute manipulation, e.g., one network to remove the mustache and another one to add it. 
The attribute manipulation is performed by a pixel-wise addition of the residual image containing the required attribute and the input image.
This approach handles a single attribute manipulation per trained model.
The IcGAN~\cite{perarnau2016icgan} is composed of an encoder and a conditional GAN generator using a normal distribution independent of the attribute to generate the latent image representation.
The input image is also encoded into an attribute information vector.
Then, the attribute manipulation is performed by modifying the attribute vector and using it and the latent representation as input to the GAN generator.
The VAE/GAN~\cite{larsen2016vaegan} generates a vector for each attribute computing the difference between the mean latent representations with and without the attribute.
Thus, the face attributes can be manipulated by adding the generated attribute vectors to a latent representation.
Also, based on an encoder/decoder network with an attribute vector, the Fader network~\cite{lample2017fader} produces a latent representation invariant to the attributes by an adversarial process introduced in the architecture.
As stated by He et al.~\cite{he2019attgan}, this process may result in information loss, which can compromise its use to our proposed attribute normalization, since some discriminant information in the periocular image can be lost.
The SaGAN model~\cite{zhang2018sagan} is composed of a generator developed with an attribute manipulation network (AMN) and a spatial attention network (SAN), and a discriminator to determine whether or not the generated image is real and for attribute classification.
The SAN and AMN models were combined in the generator to induce the manipulation only inside the attribute region.
The authors also evaluated the proposed attribute editing model on face recognition.
They used the generated images with edited attributes for data augmentation improving the verification results in two datasets.
As one can see, there are several models for facial attributes editing.
Regarding biometric system applications, it is crucial to the model the ability to modify only the desired attribute, without removing or changing any other information that may be discriminating for the subject.

\section{Proposed attribute normalization method}
\label{sec:proposed}

The proposed attribute normalization preprocess consists of applying generative deep models for image attribute editing to a pair of ocular images aiming for the correction/removal of different attributes.
Regarding the within-class variability in periocular images caused by different aspects such as eyeglasses and eye gaze, the hypothesis that we considered in this work is that it is possible to decrease this variabiality by an attribute normalization preprocess.

\begin{figure}[!ht]
 \centering
   	\includegraphics[width=\columnwidth]{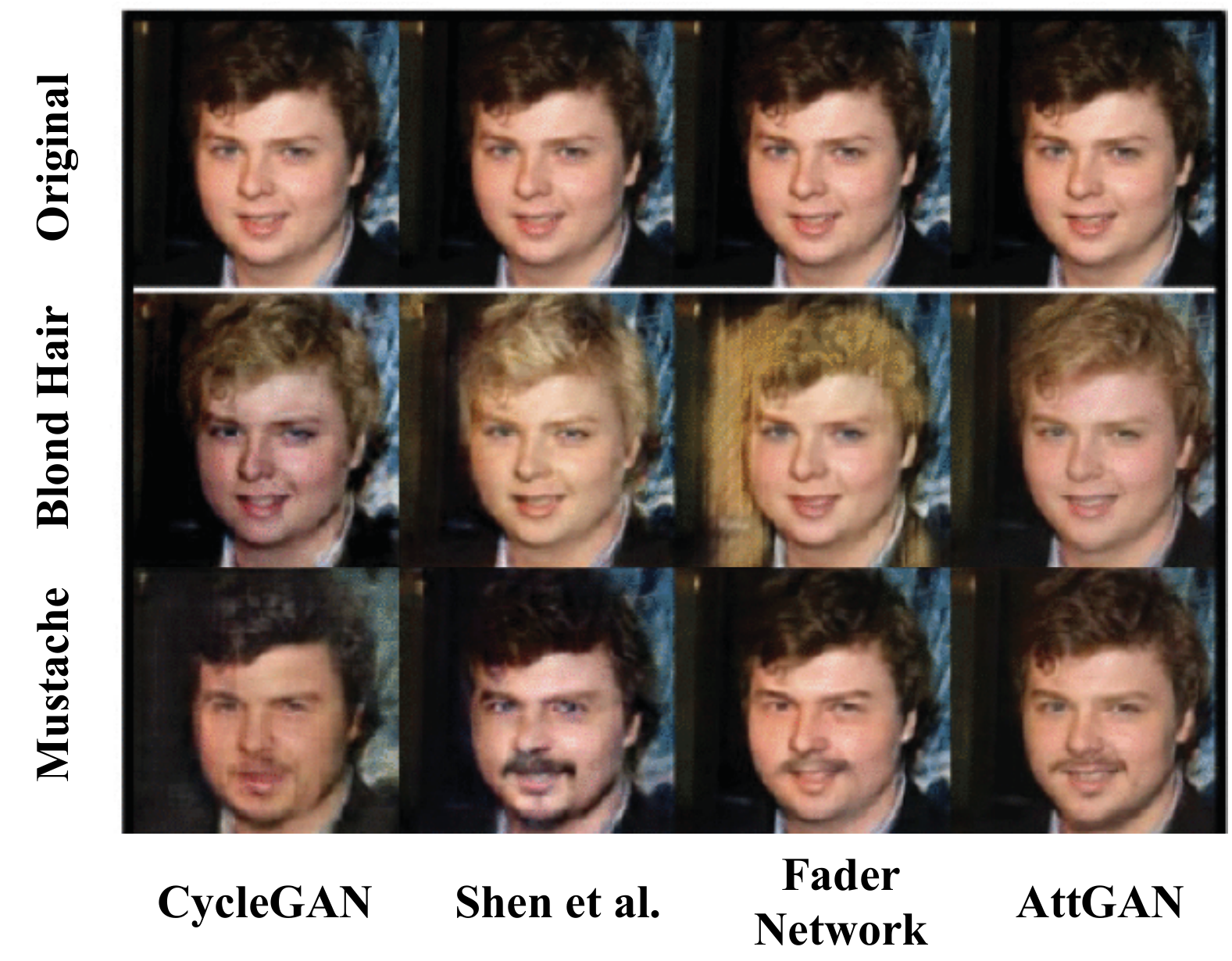}
 \caption{Comparison of state-of-the-art methods for facial attribute editing results. Adapted from~\cite{he2019attgan}.}
\label{fig:attcomp}
\end{figure}

To perform such normalization process, we employed the AttGAN model~\cite{he2019attgan} since its results compared with other state-of-the-art methods demonstrated a better capacity in changing facial attributes keeping the subject identity information as can be seen in Fig.~\ref{fig:attcomp}, which is a crucial factor for a biometric system.

The AttGAN~\cite{he2019attgan} is a deep model based on an encoder/decoder architecture.
Compared with other facial attribute editing models, its main difference is an attribute classification constraint, which requires the correct attribute manipulation in the generated images.
Regarding the problem of information loss, the architecture has a reconstruction learning, used to preserve the other attribute details, i.e., changing only the required attribute.
The model training is performed using three learning components: the reconstruction, the attribute classification, and adversarial learning.
These components guarantee the visual and reconstruction quality of the generated images with the correct attribute manipulation.
Due to all these features and mainly regarding the ability to reduce the information loss, we choose the AttGAN network to perform the proposed attribute normalization.
As the generative model receives as input an image and the attributes to be changed, we performed the attribute normalization by feeding the model with the images and requesting to remove the eyeglasses and correct the eye gaze.

The AttGAN can handle multiple attribute editing, i.e., changing more than one attribute with a single model. 
However, as we had to use different datasets for each attribute normalization in our experiments, we trained two models, one for each attribute.
We validate our proposed normalization by comparing the results of biometric systems based on handcrafted features and deep learning approaches using the original and normalized images.

\section{Experiments}
\label{sec:experiments}

\subsection{Datasets}
\label{sec:dataprot}

We carried out the experiments using two datasets: 
the UFPR-Eyeglasses, collected for this work, and used for the eyeglasses attribute normalization, and the UBIPr~\cite{padole2012ubipr} for the eye gaze normalization.
These datasets were detailed bellow.

\subsubsection{UFPR-Eyeglasses}

We collected a new challenging dataset to evaluate the effect of the occlusion caused by eyeglasses in the periocular recognition using images captured by mobile devices under real uncontrolled environments.
The dataset has $2{,}270$ periocular images (containing both eyes) from $83$ subjects ($166$ classes), all taken by the subject himself/herself using his/her smartphone at visible wavelength in $3$ distinct sessions.
We manually annotated the iris bounding box of each image, and used these annotations to perform the image normalization regarding rotation and scale, and also to crop the periocular region of each eye to $256 \times 256$ pixels.
The within-class variations are mainly caused by different aspects on the images, such as illumination, occlusions, distances, reflection, eyeglasses, and image quality.
The UFPR-Eyeglasses dataset (images and annotations) is available (under author request) to the research community at [\url{https://web.inf.ufpr.br/vri/databases/ufpr-eyeglasses/}].

\subsubsection{UBIPr}

The UBIPr dataset~\cite{padole2012ubipr} is composed of $10{,}250$ ocular images from $344$ subjects.
These images were captured under an uncontrolled environment by a Canon OS 5D camera with a $400$mm focal length at visible wavelength.
The main challenge of this dataset includes several variability factors in the images, such as different distances, scales, occlusions, poses, eye gazes, and eyeglasses.
Unlike the UFPR-eyeglasses, this dataset does not contain images from the same subject with and without eyeglasses.
Instead, there are images from the same subject with and without eye gaze.
Thus, we used this dataset to evaluate the eye gaze normalization.

\subsection{Baseline methods}
\label{sec:baseline}

We evaluated the proposed ocular normalization scheme using handcrafted features~\cite{park2011periocular, ahmed2017michewinner}, and deep representations based on approaches that recently reported state-of-the-art performances in the periocular and iris recognition~\cite{luz2018deep, zanlorensi2018impact}. These methods are detailed below.

\subsubsection{Handcrafted features approaches}
\label{sec:hand}

For the evaluation of the handcrafted features-based methods, we employed three approaches.
The first is one of the first periocular recognition methods found in the literature, proposed by Park et al.~\cite{park2011periocular}.
This approach combined Local Binary Patterns (LBP)~\cite{ojala1994lbp, ojala1996lbp}, Histogram of Oriented Gradients (HOG)~\cite{dalal2005histogram}, and Scale-Invariant Feature Transform (SIFT)~\cite{lowe2004sift} features.
The second one is the winner approach in the Miche-II contest~\cite{demarsico2017miche, ahmed2017michewinner}.
This method is also composed of an iris recognition scheme, but in our experiments, we used only the periocular recognition module, which was performed using Multi-Block Transitional Local Binary Patters (MB-TLBP) features~\cite{ahmed2017michewinner}.
At last, we combining the following features by a score-level fusion: LBP, Local Phase Quantization (LPQ)~\cite{ojansivu2008blur}, HOG and SIFT.
All the features were extracted from a gray representation of the images extracted by the intensity channel.
The normalized LBP and LPQ features were extracted from $16$ patches with a size of $64 \times 64$ pixels cropped from each image.
Then, the features of each patch were concatenated, generating feature vectors with a size of $944$ and $4096$ for the LBP and LPQ, respectively.
The HOG features were extracted from the entire image producing a feature vector with $72{,}900$ of size.

\subsubsection{Deep learning based approaches}
\label{sec:deep}

Recent works reported promising results in the development of biometric systems based on deep representations of the periocular region~\cite{luz2018deep, proenca2018prwis, diaz2018periocular, zanlorensi2019cross, diaz2019cross}.
These approaches generally consist of a CNN model that has a softmax layer at the top, and it is trained using the cross-entropy loss function.
After the training stage, the softmax layer is removed, and then the deep representations can be extracted at the newest last layer.
To evaluate the attribute normalization using these kinds of models, we employed two state-of-the-art methods to extract deep representations~\cite{luz2018deep, zanlorensi2019cross}.
These methods are based on the VGG16 and ResNet50 architectures pre-trained for face recognition~\cite{cao2017vggface2}.
Both methods generated a feature vector with a size of $256$ for each image.
We reported results from $5$ runs (repetitions) for each model.

\section{Results and Discussion}
\label{sec:results}
 
\begin{figure}[!ht]
\centering
\begin{tabular}{cc}
    \textbf{UFPR-Eyeglasses} & \textbf{UBIPr} \\

    {\includegraphics[width=0.18\columnwidth]{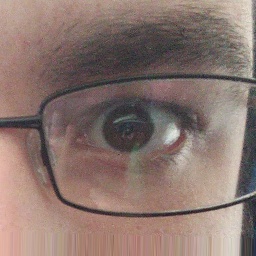}}
    {\includegraphics[width=0.18\columnwidth]{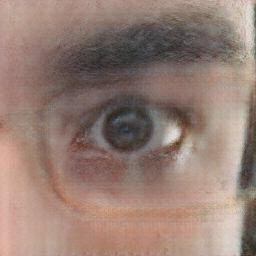}} &
    
    {\includegraphics[width=0.18\columnwidth]{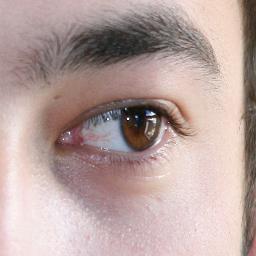}}
    {\includegraphics[width=0.18\columnwidth]{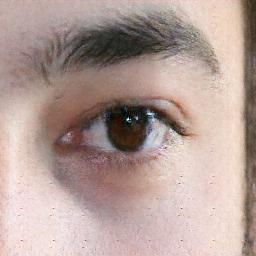}} \\
    
    {\includegraphics[width=0.18\columnwidth]{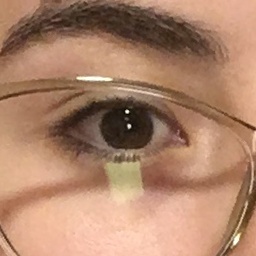}} 
    {\includegraphics[width=0.18\columnwidth]{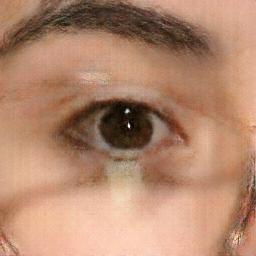}} &
    
    {\includegraphics[width=0.18\columnwidth]{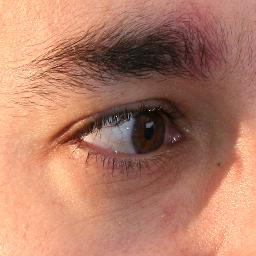}} 
    {\includegraphics[width=0.18\columnwidth]{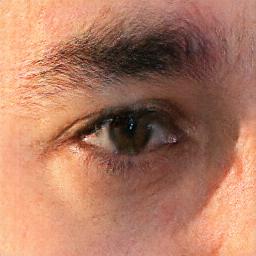}} \\

    {\includegraphics[width=0.18\columnwidth]{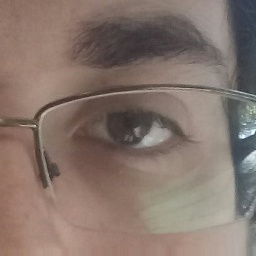}} 
    {\includegraphics[width=0.18\columnwidth]{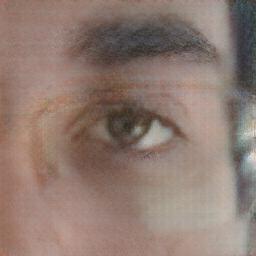}}  &
    
    {\includegraphics[width=0.18\columnwidth]{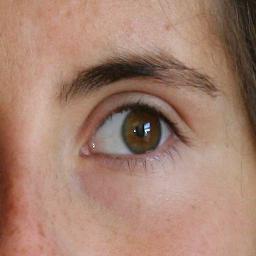}} 
    {\includegraphics[width=0.18\columnwidth]{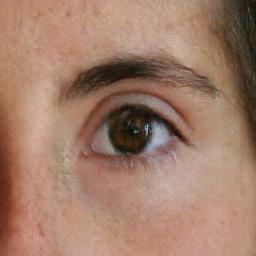}} \\
    
    \hspace{0.02\columnwidth}
    Original 
    \hspace{0.04\columnwidth}
    Normalized  &
    \hspace{0.02\columnwidth}
    Original 
    \hspace{0.04\columnwidth}
    Normalized 

\end{tabular}
\caption{Examples of original and normalized images from the UFPR-Eyeglasses (Eyeglasses removal) and UBIPr (Eyegaze correction) datasets.}
\label{fig:qualitative}
\end{figure}

The first step in our proposed normalization strategy is the training of the AttGAN model for ocular attribute editing using periocular images.
For the eyeglasses normalization (removal), we employed the entire UBIPr dataset in the training stage.
Then, we normalized all the images from the UFPR-eyeglasses dataset by removing the eyeglasses.
For the eye gaze normalization, we trained the Att-GAN using images from the first half of the subjects from the UBIPr dataset and normalized all images from the second half of the subjects by correcting the eye gaze.
The Deep learning based approaches were trained using the first half of the subjects for both datasets.
The second half of the subjects were used to evaluate and compare handcrafted features and deep learning approaches using original and normalized images.
Some qualitative results of the attribute normalization using the AttGAN model are shown in Fig.~\ref{fig:qualitative}.

For the recognition performance evaluation, according to the conclusions we previously drew about distance measures in ocular representations~\cite{luz2018deep, zanlorensi2019cross}, we chose to use the cosine distance metric to match both deep learning-based and handcrafted approaches. Regarding the SIFT features matching, we used the ratio test, as proposed by Lowe~\cite{lowe2004sift}.

We started by generating pairwise comparisons considering only images with different attributes, i.e.,  pairs with eyeglasses/no-eyeglasses in the UFPR-Eyeglasses dataset and pairs with different gaze in the case of the UBIPr dataset.
Using the second half of the subjects for each dataset, we applied the all-against-all protocol, generating $3{,}072$ genuine and $274{,}464$ impostor pairs for the UFPR-Eyeglasses dataset and $22{,}012$ genuine / $6{,}246{,}232$ impostors pairs for the UBIPr dataset. 

Considering a verification task, we used the Decidabilty index and the Area Under the Curve (AUC) as metrics to evaluate the methods.
The Decidability index measures how separated are the genuine and impostors scores distributions.
As the proposed normalization aims to decrease the within-class variability, we considered the Decidability as the primary metric.
The AUC informs the quality of the predictions based on different thresholds.
The results achieved with the proposed attribute normalization are shown in Table~\ref{tab:result}, for the UFPR-Eyeglasses and UBIPr datasets.
Note that we compared the results of the methods using the original and normalized images, in order to better evaluate the improvements in performance with respect to the solution described in this paper.

\begin{table}[!ht]
\centering
\caption{Comparison of results using original and normalized images in the UFPR-Eyeglasses and UBIPr datasets.}
\label{tab:result}
\resizebox{\columnwidth}{!}
{
\begin{tabular}{@{}lccc@{}}
\toprule 
\multirow{2}{*}{\centering{Method - Features}}                 & \multirow{2}{*}{Att. Normalization}
                                                                           & \multicolumn{2}{c}{UFPR-Eyeglasses / UBIPr}              \\ 
                                                                                  \cline{3-4}
                                                               &           & {AUC (\%)}                  & {Decidability}             \\

\midrule
\multirow{2}{*}{Ahmed et al.~\cite{ahmed2017michewinner}}      & -         & $73.0$ / $84.9$             & $0.77$ / $1.16$            \\
                                                               & Proposed  & \boldmath{$73.2$ / $85.2$}  & \boldmath{$0.79$ / $1.17$} \\
\midrule
\multirow{2}{*}{Park et al.~\cite{park2011periocular}}         & -         & $78.8$ / \boldmath{$89.6$}  & $1.11$ / \boldmath{$1.73$} \\
                                                               & Proposed  & \boldmath{$85.2$} / \unboldmath{$87.8$}  & \boldmath{$1.43$} / \unboldmath{$1.62$} \\
\midrule
LBP + LPQ +                                                    & -         & $75.9$ / \boldmath{$90.2$}  & $0.92$ / $1.71$            \\
HOG + SIFT                                                     & Proposed  & \boldmath{$87.2$} / \unboldmath{$90.0$}  & \boldmath{$1.58$ / $1.77$} \\
\midrule
\midrule
\multirow{2}{*}{Luz et al.~\cite{luz2018deep}}                 & -         & $85.9$ / \boldmath{$98.3$}  & $1.57$ / \boldmath{$3.64$} \\
                                                               & Proposed  & \boldmath{$89.0$} / \unboldmath{$98.1$}  & \boldmath{$1.81$} / \unboldmath{$3.50$} \\
\midrule
\multirow{2}{*}{Zanlorensi et al.~\cite{zanlorensi2019cross}}  & -         & $92.2$ / $99.2$             & $2.09$ / $4.00$            \\
                                                               & Proposed  & \boldmath{$92.9$ / $99.4$}  & \boldmath{$2.16$ / $4.14$} \\
\bottomrule
\end{tabular}
}
\end{table}

\begin{figure}[!tb]
\centering
\setlength{\tabcolsep}{1.5pt}
\begin{tabular}{cccc}
    \textbf{Original} & \textbf{Normalized} & \textbf{Original} & \textbf{Normalized}\\

    {\includegraphics[width=0.11\columnwidth]{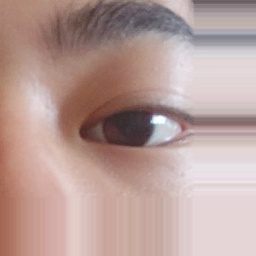}}
    {\includegraphics[width=0.11\columnwidth]{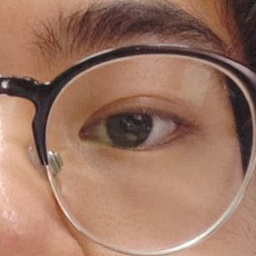}}&
    
    {\includegraphics[width=0.11\columnwidth]{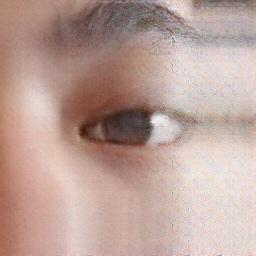}}
    {\includegraphics[width=0.11\columnwidth]{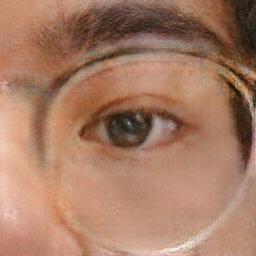}}&
    
    {\includegraphics[width=0.11\columnwidth]{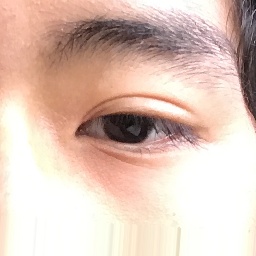}}
    {\includegraphics[width=0.11\columnwidth]{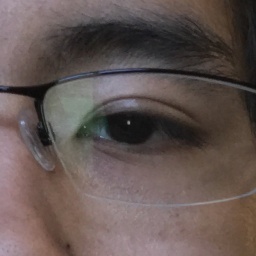}}& 
    
    {\includegraphics[width=0.11\columnwidth]{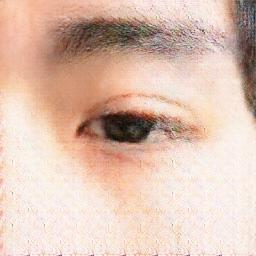}} 
    {\includegraphics[width=0.11\columnwidth]{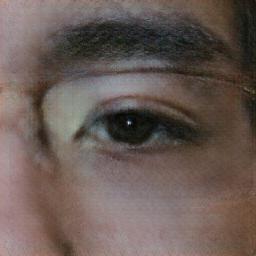}} \\
    
     $0.24$ & $0.87$ & $0.25$ & $0.89$\\

    {\includegraphics[width=0.11\columnwidth]{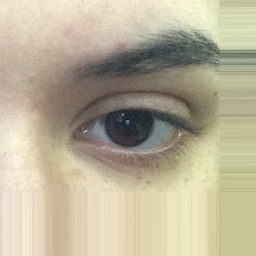}}
    {\includegraphics[width=0.11\columnwidth]{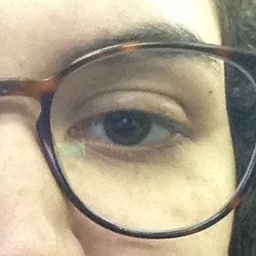}}&
    
    {\includegraphics[width=0.11\columnwidth]{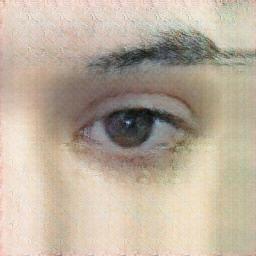}}
    {\includegraphics[width=0.11\columnwidth]{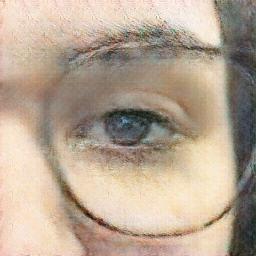}}&
    
    {\includegraphics[width=0.11\columnwidth]{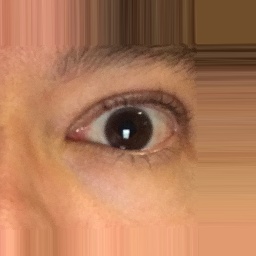}}
    {\includegraphics[width=0.11\columnwidth]{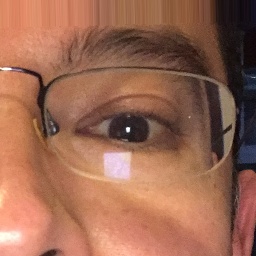}}& 
    
    {\includegraphics[width=0.11\columnwidth]{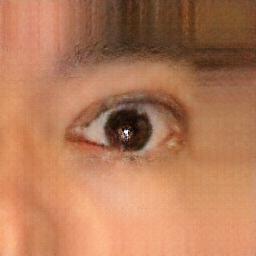}} 
    {\includegraphics[width=0.11\columnwidth]{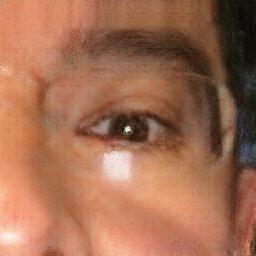}} \\

    $0.39$ & $0.64$ & $0.40$ & $0.92$\\
    
    {\includegraphics[width=0.11\columnwidth]{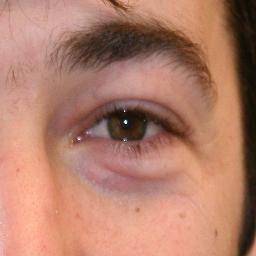}}
    {\includegraphics[width=0.11\columnwidth]{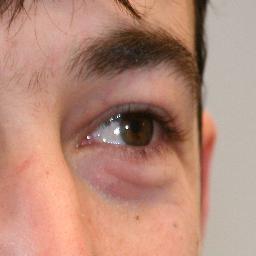}}&
    
    {\includegraphics[width=0.11\columnwidth]{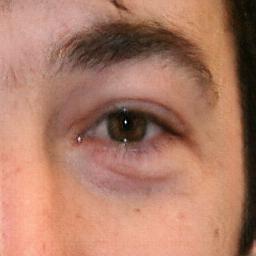}}
    {\includegraphics[width=0.11\columnwidth]{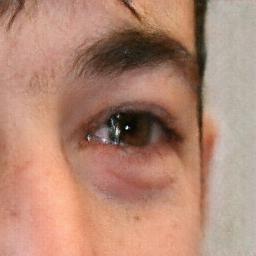}}&
    
    {\includegraphics[width=0.11\columnwidth]{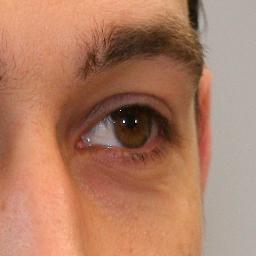}}
    {\includegraphics[width=0.11\columnwidth]{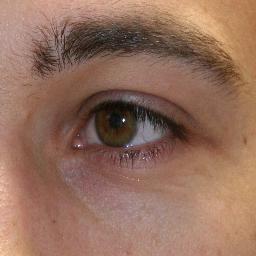}}& 
    
    {\includegraphics[width=0.11\columnwidth]{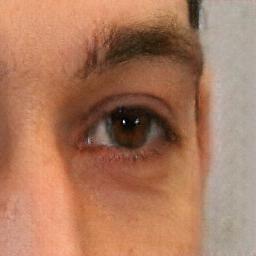}}
    {\includegraphics[width=0.11\columnwidth]{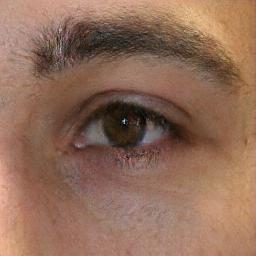}} \\

    $0.58$ & $0.92$ & $0.59$ & $0.90$\\
    
    {\includegraphics[width=0.11\columnwidth]{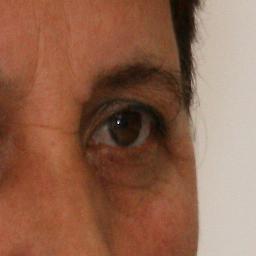}}
    {\includegraphics[width=0.11\columnwidth]{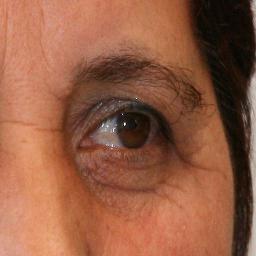}}&
    
    {\includegraphics[width=0.11\columnwidth]{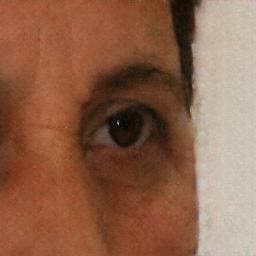}}
    {\includegraphics[width=0.11\columnwidth]{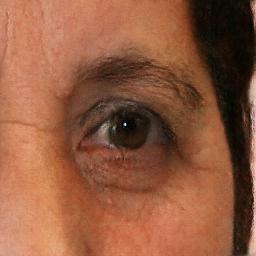}}&
    
    {\includegraphics[width=0.11\columnwidth]{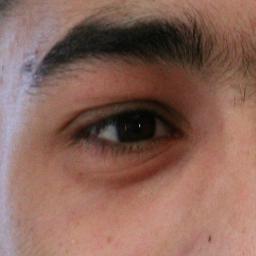}}
    {\includegraphics[width=0.11\columnwidth]{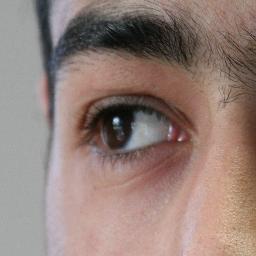}}& 
    
    {\includegraphics[width=0.11\columnwidth]{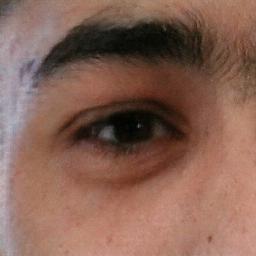}}
    {\includegraphics[width=0.11\columnwidth]{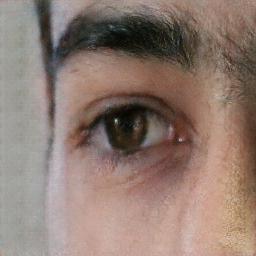}} \\

    $0.56$ & $0.91$ & $0.66$ & $0.92$\\

\end{tabular}
\caption{Genuine scores comparison from original and normalized images. Higher scores mean that the periocular image pairwise is more likely to be genuine.}
\label{fig:gencomp}
\end{figure}

The results showed that the proposed normalization preprocessing consistently improve the verification results in the UFPR-Eyeglasses dataset, increasing the Decidability by $28\%$ (i.e., $1.4261/1.1093$) and $71\%$ (i.e., $1.5764/0.9206$), respectively using the features from the method proposed by Park et al.~\cite{park2011periocular} and from the proposed handcrafted features fusion.
Using the deep learning based approaches, the attribute normalization improved the Decidability by $15\%$ and $4\%$ for the methods proposed by Luz et al.~\cite{luz2018deep} and Zanlorensi et al.~\cite{zanlorensi2019cross}, respectively.
Unlike the experiments performed using the UFPR-Eyeglasses dataset, in the UBIPr one, the attribute normalization process consists of the eye gaze correction.
Since this process is computed in a small portion of the periocular image (only in the eyeball region), in general, we can observe that the impact of applying the attribute normalization is smaller than the ones obtained in the UFPR-Eyeglasses images.
Nevertheless, the highest Decidability index in the UBIPr dataset using hand-crafted features and Deep learning-based models was achieved by employing the normalized images.

Fig.~\ref{fig:gencomp} shows some qualitative results where wrong genuine matching between original images were corrected using the proposed attribute normalization. One can also observe that in the UFPR-Eyeglasses dataset, even when the eyeglasses were not entirely removed, the generative model was able to smooth them, such that the biometric system was able to correctly classified a pair as genuine.
Investigating other wrong genuine matches, we stated that the pose and illumination aspect is one of the most significant factors that penalize the within-class variability in the UBIPr dataset.

\section{Conclusion}
\label{sec:conclusion}

This paper proposed an attribute normalization scheme that can be used as a preprocessing step to reduce the within-class variability in unconstrained periocular recognition. The idea is to use state-of-the-art generative model that normalizes specific factors of all samples before being used by the recognition algorithm.
Noting that our solution is fully agnostic to the recognition method used, our proof-of-concept was conducted in two datasets and five different baseline methods. 
Our idea was to compare the levels of performance attained by the recognition methods when using the raw data and when receiving the images preprocessed by our solution. 
The observed results corroborated our hypothesis that the proposed attribute normalization is highly effective to reduce the within-class variabilities, without compromising the discriminability between classes, which is the root for the observed improvements in performance.

\vfill
\noindent
\textbf{Acknowledgment:} This work was supported by grants from the National Council for Scientific and Technological Development (CNPq)(\#313423/2017-2 and \#428333/2016-8), and the Coordination for the Improvement of Higher Education Personnel (CAPES), Brazilian funding agencies, and also gratefully acknowledge the support of NVIDIA Corporation with the donation of the Titan Xp GPU used for this research.
The second author work is funded by FCT/MEC through national funds and co-funded by FEDER - PT2020 partnership agreement under the projects UID/EEA/50008/2019 and POCI-01-0247-FEDER-033395.

%\footnotesize
%\scriptsize
\balance
\setlength{\bibsep}{3pt}
\bibliographystyle{IEEEbib}
\bibliography{paper}

\begin{thebibliography}{10}

\bibitem{luz2018deep}
E.~{Luz}, G.~{Moreira}, L.~A. {Zanlorensi Junior}, and D.~{Menotti},
\newblock ``Deep periocular representation aiming video surveillance,''
\newblock {\em Pattern Recognition Letters}, vol. 114, pp. 2--12, 2018.

\bibitem{proenca2018prwis}
H.~{Proença} and J.~C. {Neves},
\newblock ``{Deep-PRWIS}: Periocular recognition without the iris and sclera
  using deep learning frameworks,''
\newblock {\em IEEE Transactions on Information Forensics and Security}, vol.
  13, no. 4, pp. 888--896, Apr 2018.

\bibitem{zhao2018critical}
Z.~{Zhao} and A.~{Kumar},
\newblock ``Improving periocular recognition by explicit attention to critical
  regions in deep neural network,''
\newblock {\em IEEE Transactions on Information Forensics and Security}, vol.
  13, no. 12, pp. 2937--2952, Dec 2018.

\bibitem{diaz2018periocular}
K.~{Hernandez-Diaz}, F.~{Alonso-Fernandez}, and J.~{Bigun},
\newblock ``Periocular recognition using cnn features off-the-shelf,''
\newblock in {\em BIOSIG}, Sep. 2018, pp. 1--5.

\bibitem{zanlorensi2019cross}
L.~A. {Zanlorensi}, D.~R. {Lucio}, A.~S. {Britto Jr.}, H.~{Proença}, and
  D.~{Menotti},
\newblock ``Deep representations for cross-spectral ocular biometrics,''
\newblock {\em IET Biometrics}, November 2019.

\bibitem{diaz2019cross}
F.~{Alonso-Fernandez} K.~{Hernandez-Diaz} and J.~{Bigun},
\newblock ``Cross spectral periocular matching using resnet features,''
\newblock in {\em ICB}, 2019, pp. 1--6.

\bibitem{sequeira2017cross}
A.~F. {Sequeira} et~al.,
\newblock ``Cross-eyed 2017: Cross-spectral iris/periocular recognition
  competition,''
\newblock in {\em IJCB}, Denver, CO, USA, Oct 2017, pp. 725--732, IEEE.

\bibitem{demarsico2017miche}
M.~{De Marsico}, M.~{Nappi}, and H.~{Proença},
\newblock ``{Results from MICHE II - Mobile Iris CHallenge Evaluation II},''
\newblock {\em Pattern Recognition Letters}, vol. 91, pp. 3--10, may 2017.

\bibitem{ahmed2017michewinner}
N.~U. {Ahmed}, S.~{Cvetkovic}, E.~H. {Siddiqi}, A.~{Nikiforov}, and
  I.~{Nikiforov},
\newblock ``Combining iris and periocular biometric for matching visible
  spectrum eye images,''
\newblock {\em Pattern Recognition Letters}, vol. 91, pp. 11--16, may 2017.

\bibitem{zanlorensi2019survey}
L.~A. {Zanlorensi}, R.~{Laroca}, E.~{Luz}, A.~S. {Britto Jr.}, L.~S.
  {Oliveira}, and D.~{Menotti},
\newblock ``Ocular recognition databases and competitions: A survey,''
\newblock {\em arXiv preprint}, vol. arXiv:1911.09646, pp. 1--20, 2019.

\bibitem{goodfellow2014gan}
I.~{Goodfellow} et~al.,
\newblock ``Generative adversarial nets,''
\newblock in {\em Advances in Neural Information Processing Systems 27}, pp.
  2672--2680. Curran Associates, Inc., 2014.

\bibitem{kingma2014vae}
D.~P. {Kingma} and M.~{Welling},
\newblock ``Auto-encoding variational bayes,''
\newblock in {\em International Conference on Learning Representations}, 2014.

\bibitem{larsen2016vaegan}
A.~B.~L. {Larsen}, S.~K. {Sønderby}, H.~{Larochelle}, and O.~{Winther},
\newblock ``Autoencoding beyond pixels using a learned similarity metric,''
\newblock in {\em ICML}, New York, New York, USA, Jun 2016, vol.~48, pp.
  1558--1566, PMLR.

\bibitem{perarnau2016icgan}
G.~{Perarnau}, Joost {van de Weijer}, Bogdan {Raducanu}, and Jose~M.
  {\'Alvarez},
\newblock ``{Invertible Conditional GANs for image editing},''
\newblock in {\em NIPS Workshop on Adversarial Training}, 2016.

\bibitem{choi2018stargan}
Y.~{Choi} et~al.,
\newblock ``Stargan: Unified generative adversarial networks for multi-domain
  image-to-image translation,''
\newblock in {\em CVPR}, June 2018.

\bibitem{lample2017fader}
G.~{Lample} et~al.,
\newblock ``Fader networks:manipulating images by sliding attributes,''
\newblock in {\em Advances in Neural Information Processing Systems}, pp.
  5967--5976. Curran Associates, Inc., 2017.

\bibitem{he2019attgan}
Z.~{He}, W.~{Zuo}, M.~{Kan}, S.~{Shan}, and X.~{Chen},
\newblock ``Attgan: Facial attribute editing by only changing what you want,''
\newblock {\em IEEE Transactions on Image Processing}, vol. 28, no. 11, pp.
  5464--5478, Nov 2019.

\bibitem{shen2017resface}
W.~{Shen} and R.~{Liu},
\newblock ``Learning residual images for face attribute manipulation,''
\newblock in {\em CVPR}, July 2017.

\bibitem{zhu2017cyclegan}
J.~{Zhu}, T.~{Park}, P.~{Isola}, and A.~A. {Efros},
\newblock ``Unpaired image-to-image translation using cycle-consistent
  adversarial networks,''
\newblock in {\em ICCV}, Oct 2017.

\bibitem{karras2019stylegan}
T.~{Karras}, S.~{Laine}, and T.~{Aila},
\newblock ``A style-based generator architecture for generative adversarial
  networks,''
\newblock in {\em CVPR}, June 2019.

\bibitem{xiao2018elegant}
T.~{Xiao}, J.~{Hong}, and J.~{Ma},
\newblock ``Elegant: Exchanging latent encodings with gan for transferring
  multiple face attributes,''
\newblock in {\em ECCV}, September 2018.

\bibitem{zhang2018sagan}
Gang Zhang, Meina Kan, Shiguang Shan, and Xilin Chen,
\newblock ``Generative adversarial network with spatial attention for face
  attribute editing,''
\newblock in {\em ECCV}, September 2018.

\bibitem{padole2012ubipr}
C.~N. {Padole} and H.~{Proença},
\newblock ``Periocular recognition: Analysis of performance degradation
  factors,''
\newblock in {\em International Conference on Biometrics (ICB)}, March 2012,
  pp. 439--445.

\bibitem{park2011periocular}
U.~{Park}, R.~R. {Jillela}, A.~{Ross}, and A.~K. {Jain},
\newblock ``Periocular biometrics in the visible spectrum,''
\newblock {\em IEEE Transactions on Information Forensics and Security}, vol.
  6, no. 1, pp. 96--106, March 2011.

\bibitem{zanlorensi2018impact}
L.~A. {Zanlorensi}, E.~{Luz}, R.~{Laroca}, A.~S. {Britto Jr.}, L.~S.
  {Oliveira}, and D.~{Menotti},
\newblock ``The impact of preprocessing on deep representations for iris
  recognition on unconstrained environments,''
\newblock in {\em Conference on Graphics, Patterns and Images (SIBGRAPI)}. Oct
  2018, pp. 289--296, IEEE.

\bibitem{ojala1994lbp}
T.~{Ojala}, M.~{Pietikainen}, and D.~{Harwood},
\newblock ``Performance evaluation of texture measures with classification
  based on kullback discrimination of distributions,''
\newblock in {\em ICPR}. IEEE, 1994, vol.~1, pp. 582--585.

\bibitem{ojala1996lbp}
T.~{Ojala}, M.~{Pietikäinen}, and D.~{Harwood},
\newblock ``A comparative study of texture measures with classification based
  on featured distributions,''
\newblock {\em Pattern Recognition}, vol. 29, no. 1, pp. 51--59, 1996.

\bibitem{dalal2005histogram}
N.~{Dalal} and B.~{Triggs},
\newblock ``Histograms of oriented gradients for human detection,''
\newblock in {\em CVPR}, June 2005, vol.~1, pp. 886--893.

\bibitem{lowe2004sift}
D.~G. {Lowe},
\newblock ``Distinctive image features from scale-invariant keypoints,''
\newblock {\em International Journal of Computer Vision}, vol. 60, pp. 91--110,
  2004.

\bibitem{ojansivu2008blur}
V.~{Ojansivu} and J.~{Heikkil{\"a}},
\newblock ``Blur insensitive texture classification using local phase
  quantization,''
\newblock in {\em International conference on image and signal processing}.
  Springer, 2008, pp. 236--243.

\bibitem{cao2017vggface2}
Q.~{Cao}, L.~{Shen}, W.~{Xie}, O.~M. {Parkhi}, and A.~{Zisserman},
\newblock ``{VGGFace2}: {A} dataset for recognising faces across pose and
  age,''
\newblock {\em CoRR}, 2017.

\end{thebibliography}

\end{document}